\pdfoutput=1

\documentclass[conference]{IEEEtran}

\ifCLASSINFOpdf
\else
\fi
\usepackage{amsmath, amssymb, bm}
\usepackage{caption} 
\usepackage{booktabs}
\usepackage{enumitem} 
\usepackage{graphicx}
\usepackage{algorithm}
\usepackage{algcompatible}
\usepackage{algpseudocode}
\usepackage{amsfonts}
\usepackage{dsfont}
\usepackage[makeroom]{cancel}
\usepackage{tabularx}
\usepackage{xcolor}
\usepackage{bbm}

\usepackage[T1]{fontenc}
\usepackage[font=small,labelfont=bf,tableposition=top]{caption}

\usepackage{capt-of}

\usepackage{accents}

\usepackage{algcompatible}
\usepackage{algpseudocode}
\usepackage{pifont}

\PassOptionsToPackage{hyphens}{url}\usepackage{hyperref}

\usepackage{paracol}
\usepackage{xparse}
\usepackage{kantlipsum}

\NewDocumentCommand\myframedtext{ s O{.9\linewidth} m }{%
    \IfBooleanTF{#1}{\begin{figure*}}{\begin{figure}}%
      \centering%
      \fbox{\parbox{#2}{%
        #3%
      }}%
    \IfBooleanTF{#1}{\end{figure*}}{\end{figure}}}
    
\newcommand{\ds}{\displaystyle}
\newcommand{\df}{\displaystyle\frac}

\newcommand{\quotes}[1]{``#1''}

\newcommand{\cond}{\: | \:}
\newcommand{\one}[1]{\mathds{1}_{#1}}
\newcommand{\E}{\mathbb{E}}

\makeatletter
\newcommand*\bigcdot{\mathpalette\bigcdot@{.5}}
\newcommand*\bigcdot@[2]{\mathbin{\vcenter{\hbox{\scalebox{#2}{$\m@th#1\bullet$}}}}}
\makeatother

\newcommand{\name}{\ensuremath{\mathrm{PROPS} \;}}

\newcommand{\bbtext}{source \,}  
\newcommand{\lptext}{perturbation \,}  
\newcommand{\sourcetext}{\bbtext}  
\newcommand{\perturbationtext}{\lptext}  

\newcommand{\bbtextabbrev}{s}  
\newcommand{\lptextabbrev}{p}  
\newcommand{\rnn}{\bbtext}  

\newcommand{\generic}{\mathcal{K}_{\bbtextabbrev}\:} 
\newcommand{\pretrained}{\generic}

\newcommand{\blackbox}{\generic}
\newcommand{\source}{\generic}

\newcommand{\perturbed}{\mathcal{K}_{\lptextabbrev}\:} 
\newcommand{\perturbation}{\perturbed} 

\newcommand{\numbb}{K_{\bbtextabbrev}}
\newcommand{\numlp}{K_{\lptextabbrev}}

\begin{document}

\title{PROPS: Probabilistic personalization of black-box sequence models}

\author{\IEEEauthorblockN{Michael Thomas Wojnowicz and Xuan Zhao}
\IEEEauthorblockA{Department of Research and Intelligence \\
Cylance, Inc. \\
Irvine, California 92612 \\
\{mwojnowicz, xzhao\}@cylance.com}}

\maketitle

\begin{abstract}

We present \name, a lightweight transfer learning mechanism for sequential data. \name learns probabilistic perturbations around the predictions of one or more arbitrarily complex, pre-trained black box models (such as recurrent neural networks).   The technique pins the black-box prediction functions to \quotes{source nodes} of a hidden Markov model (HMM), and uses the remaining nodes as \quotes{perturbation nodes} for learning customized perturbations around those predictions. In this paper, we describe the \name model, provide an algorithm for online learning of its parameters, and demonstrate the consistency of this estimation.   We also explore the utility of \name in the context of personalized language modeling.   In particular, we construct a baseline language model by training a LSTM on the entire Wikipedia corpus of 2.5 million articles (around 6.6 billion words), and then use \name to provide lightweight customization into a personalized language model of President Donald J. Trump's tweeting. 
We achieved good customization after only 2,000 additional words, and find that the \name model, being fully probabilistic, provides insight into when President Trump's speech departs from generic patterns in the Wikipedia corpus.    Python code (for both the \name training algorithm as well as experiment reproducibility) is available at \url{https://github.com/cylance/perturbed-sequence-model}.  

\end{abstract}

\IEEEpeerreviewmaketitle


\section{Introduction}

%

Suppose one has access to one, or possibly more, pre-trained sequence models $\{\mathcal{M}_k\}$ which have been trained in one context, but whose knowledge should be transferred to a related context.   Further suppose that one requires the transfer learning to be lightweight and streaming, and that after training, one would like to have a {\bf fully probabilistic sequence model} (i.e one with stochastic hidden variables).    The fully probabilistic sequence model would provide real-time probabilistic statements about membership in the previous, reference context versus membership in the new, customized context.   

One motivating example is a situation where one can train a single, arbitrarily complex, black-box sequence model in the cloud on a large public corpus, but then would like to \quotes{customize} or \quotes{personalize} that model to a particular user's private data on an endpoint (laptop, mobile device, etc.).   One can imagine this scenario at play for machine learning models for automatic message generation or text-to-speech.   In such contexts, a lightweight, streaming transfer learning mechanism would minimize computation and memory footprint at deployment.  At the same time, the ability to do inference on hidden state variables would allow one to 'draw attention' to subsequences which deviate from the population-level \quotes{normal} behavior, even if they are normal for that particular user. 

To solve this problem, we present  $\name$, an acronym derived from \emph{PRObabilistically Perturbed Sequence modeling}.   
\name generalizes a streaming training algorithm for a HMM to learn customized perturbations around the predictions of pre-trained black-box sequence models.      In particular, the \name model takes $\numbb$ pre-trained black-box sequence models and pins them at $\numbb$ nodes of a HMM wrapper.    By \quotes{pinning}, we mean that the emissions distributions at those $\numbb$ {\bf source states} are not updated during training.    These source states can provide knowledge from a reference or baseline corpus about arbitrarily complex functional relationships between the sequence observed so far and upcoming observations, because the $\{\mathcal{M}_k\}$'s could be recurrent neural networks (RNNs), such as Long Short-Term Memory (LSTM) networks \cite{hochreiter}, variations thereof, or anything desired.   Indeed, the only constraint is that each such model produces predictive output probabilities, $P(y_{t+1} \cond y_{1:t})$ at each time step.

\begin{figure}[htp]
  \includegraphics[width=0.5\textwidth]{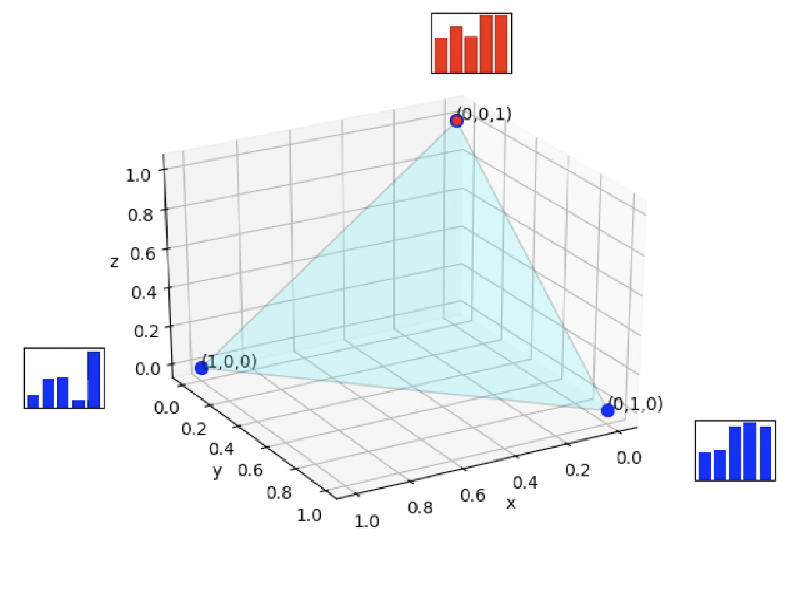}
  \caption{\emph{A probability simplex supporting the latent state distribution of a \name model with 3 latent states}}
  \label{img:PROPS}
\end{figure}

The remaining $\numlp$ nodes serve as the {\bf perturbation states}.  For instance, Figure \ref{img:PROPS} shows a probability simplex supporting the latent state distribution of a \name model with 3 latent states.     Suppose the top node (colored red) is a single source node, whereas the bottom nodes (colored blue) are the perturbation nodes.   The training of \name causes the blue nodes to converge to stable emission distributions (just as in a standard HMM), even while the red node is tied to a stochastic process of distributions that varies (potentially wildly) from timestep to timestep.  Moreover, the model learns transition dynamics across the full heterogeneous system.   
 
 



\subsection{Summary of contributions}

The \name model is a lightweight transfer learning mechanism for sequence data. \name  learns probabilistic perturbations around the predictions of an arbitrarily complex, pre-trained black box model trained on a reference corpus.   The key properties of the \name model and training algorithm are:  

\begin{itemize}
\item {\bf Streaming} - Both training and scoring can be done in a streaming (i.e. online) manner.   Among other things, this reduces memory footprint, and keeps the model's scores up-to-date with potential behavioral non-stationaries. 
\item {\bf Lightweight} - The model can be trained reliably (from being high bias/low variance), has a low computational footprint, and offers quick training times (e.g. see  \cite{panzner}, with additional speedups due to the streaming algorithm \cite{cappe}). 
\item {\bf Probabilistic/Interpretable} - The model is fully probabilistic.  In particular, its hidden states are random variables.  Thus inference -- such as filtering, smoothing, and MAP estimates -- can be performed on the hidden state variables. Because the latent states can be partitioned into nodes related to the original task and nodes related to the new task, we can make probabilistic statements about when behavior drifts from the source or reference context into the customization or transfer context.   (Note that anomalies for the \name model would be behavior that departs from both contexts, or which moves between contexts in unexpected ways.)
\item {\bf Privacy-Preserving} - Training data can be kept private while \name provides customization.   
\end{itemize}

In this paper, we describe the \name model, provide an algorithm for fitting it, corroborate the consistency of its estimators, and explore its utility in the context of personalized language modeling.

\section{Models}

\subsection{Hidden Markov Model}

A hidden Markov model (HMM) is a bivariate stochastic process $(X_t, Y_t)_{t \geq 0}$ where:
\begin{enumerate}
\item $X=(X_t)_t$ is an unobserved ergodic Markov chain (often called a hidden state sequence) over $\{ 1, \hdots, K\}$ with transition matrix $\tau$ and initial state distribution $\pi$.
\item $Y=(Y_t)_t$ is observed data, independent conditional on $X$ with parametric emission distribution $(Y_t \cond X_t =k) \sim \epsilon^{\lambda_k}$.  
\end{enumerate}

The complete data likelihood for a HMM is 
\begin{equation}
\label{eq:complete_data_likelihood}
\mbox{\scriptsize$  P(X,Y \cond \theta) = P(X_1 \cond \theta ) P(Y_1 \cond X_1, \theta ) \ds\prod_{t=2}^T P(X_t \cond X_{t-1},  \theta) P(Y_t \cond X_t, \theta) $}
\end{equation}

where parameter $\theta =(\pi, \tau, (\lambda_k)_k)$.


\subsection{\name Model}

A probabilistically perturbed sequence (\name) model is a generalization of HMM where deterministic predictive probability functions $(f^k_{Y_{1:t-1}}(Y_t))_k$ are supplied as inputs and
\begin{align*}
 (Y_t =y_t \cond Y_{1:t-1},  X_t =k) \sim   \hphantom{need to take up some space} \\
\begin{cases}
\epsilon^{\lambda_k^t}(y_t)= f^k_{Y_{1:t-1}}(y_t) & k \in \generic \\
\epsilon^{\lambda_k}(y_t) & k \in \perturbed \\ 
\end{cases} 
\label{eq:model}
\end{align*}
for partition $\generic \sqcup \perturbed =\mathcal{K} = \{1, \hdots, K \}.$

That is, a subset of states have emissions distributions that break the conditional independence assumption on the observations $Y$.  However, these emissions distributions are not learned, but provided as fixed functions.    This results in time-dependent (but unlearned) emission distribution parameters $\lambda_k^t$ governing the distribution of $(Y_t \cond X_t =k)$ for a subset of states.   Emissions distributions for the remaining states are standard, and serve to learn customized perturbations around the deterministic predictive probability functions.  Note that a streaming HMM is a special case of the \name model, with $K_{\bbtextabbrev} = 0$. 

\section{Notation}

Notation is summarized in Table~\ref{table:notation}.   Learned \name model parameters are given by $\theta=\big(\pi, \tau, \{\lambda_k \}_{k \in \perturbed} \big)$.  For simplicity of presentation (and due to the streaming context), we assume the initial state distribution $\pi$ to be fixed.    The non-learned model objects are given by $(K, \numbb, \{ f^k_{\cdot}(\cdot)\}_{k \in \blackbox})$.     


\begin{table}[ht]
\caption{Summary of notation}
\centering 
\small
\fontsize{8}{8.5}\selectfont
\setlength{\tabcolsep}{6pt} 
\label{table:notation}
\def\arraystretch{1.28}
\begin{tabularx}{0.8\linewidth}{@{}rX@{}} 
\toprule
 $\{\mathcal{M}_k\}_{k \in \generic}$ & \sourcetext models\\
 $\{f^k_{y_{1:t}}(\cdot)\}_{k \in \generic}$ & Predictive probability functions from \sourcetext models\\
$y$ & observed sequence\\
$y_t$ & observation at timestep $t$\\
$x$ & hidden state sequence\\
$x_t$ & hidden state at timestep $t$ \\
$T$ & number of timesteps \\
$W$ & number of ``words`` for discrete emissions  \\
$K_{\bbtextabbrev}, K_{\lptextabbrev}, K$ & \# \sourcetext states, \# \perturbationtext states, and total \# hidden states \\
$\generic,\perturbed, \mathcal{K}$ &  \sourcetext  states, \perturbationtext  states, hidden states \\
$\theta$ & learned \name model parameters \\
$\pi$ & initial state distribution \\
$\tau $ & state transition matrix \\
 $\{\epsilon_k \}_{k \in \mathcal{K}}$ & emissions distributions \\
$ s(X_{t-1}, X_t, Y_t)$ & complete-data sufficient statistics \\ 
\bottomrule
\end{tabularx}
\end{table}

\section{Streaming training for Hidden Markov Models}  

Here we present material on streaming (also called online, recursive, or adaptive) learning of fixed model parameters in HMMs.   The purpose is two-fold:  (1)  Algorithm \ref{alg} requires it for implementation, and (2) the overview is helpful to motivate the argument that \name is consistent. 

\subsection{Classical (batch) Estimation}

Estimating (or fitting) the model by maximum likelihood means finding the parameters $\arg \max_\theta P(x,y \cond \theta)$.   Yet because the HMM has latent variables, the sufficient statistics for the complete-data log likelihood are not observed. 
The {\bf Expectation Maximization} (EM) algorithm handles this by iteratively computing expected sufficient statistics for the latent variables (given the current parameter estimates) before maximizing the parameters.  That is, we iteratively compute  
\begin{equation}
\label{eq:em} 
\arg \max_\theta  \E_{p(x \cond y, \theta^i)} \log p(x,y \cond \theta) \quad \quad \text{for} \; \; i=1, 2 \hdots 
\end{equation} 
until convergence.   In particular, on the $i$th iteration, we can do
\begin{enumerate}
\item \emph{E-step}: Compute the {\bf expected complete-data sufficient statistics} (ESS), which for a HMM is
\begin{equation}
\label{eq:ESS}
S^{i+1} = \df{1}{T} \: \mathbb{E}_{\nu, \theta^i} \big[ \ds\sum_{t=0}^T s(X_{t-1}, X_t, Y_t) \cond Y_{0:T} ]  
\end{equation}
where $s(\cdot)$ are the sufficient statistics from the complete-data likelihood.\footnote{For example, in the case of discrete emissions, $s(X_{t-1}, X_t, Y_t) = \mathds{1}_{(X_{t-1}=i, X_t =j, Y_t = w)}$.}

\item \emph{M-step}:  Update the parameter estimate to $\theta_{i+1}  = \overline{\theta}(S_{i+1})$, where $\overline{\theta}$ refers to the operation of finding the maximum likelihood estimator. 
\end{enumerate}


\subsection{Streaming Estimation}

Classical (batch) estimation requires multiple passes through the entire data set -- one pass for each iteration of the EM algorithm.  
In the streaming (or online) setting, the goal is  to see an observation once, update model parameters, and discard the observation forever.    We follow the procedure of~\cite{cappe}, which essentially does streaming E-steps based on the data seen so far, and then a partial maximization for the M-step (i.e. model parameters are optimized prematurely, before strictly completing the E-step pass).  Compared to batch learning, this technique takes less training time to get to the same level of estimation accuracy.

\subsubsection{Overview}

Here we compute \emph{streaming estimates} of the expected complete-data sufficient statistics
\begin{equation}
\label{ESS_current}
S^{t} = \df{1}{t} \mathbb{E}_{\nu, \theta^t} \big[ \ds\sum_{r=0}^t s(X_{r-1}, X_r, Y_r) \cond Y_{0:t} ]  
\end{equation}

Streaming updates can be made to this quantity by decomposing it into two simpler functions.  In addition to the typical {\bf filter} function,
\begin{equation}
\label{filter}
\phi_{t, \nu, \theta}(k)= P_{\nu, \theta}(X_t =k \cond Y_{0:t}) \\
\end{equation}
we also define an {\bf auxilliary function} as
\begin{equation}
\label{auxiliary_function}
\rho_{t, \nu, \theta}(i,k ; \theta)= \df{1}{t} \mathbb{E}_{\nu, \theta} \big[ \ds\sum_{r=0}^t s(X_{r-1}, X_r, Y_r) \cond Y_{0:t}, X_t=k \big]  \\
\end{equation}

Applying the \emph{law of iterated expectation} with the filter (\ref{filter}) and auxiliary function (\ref{auxiliary_function}), we can compute the \emph{currently estimated expected complete-data sufficient statistics} as
\begin{equation}
\label{estimand_decomposed}
S_{t}  = \ds\sum_x{ \widehat{\phi}_{t,\nu,\theta}(x) \widehat{\rho}_{t,\nu,\theta}(x)}
\end{equation}

\subsubsection{Filter recursion}


The filter recursion can be derived using elementary probability laws:
\begin{align}
& \mbox{\scriptsize$ \phi_{t+1}(x)  \propto P(X_{t+1}=x , Y_{0:t+1}) $} \nonumber \\
&  \mbox{\scriptsize$ = P(Y_{t+1} \cond X_{t+1}=x) P(X_{t+1}=x, Y_{0:t}) $}  \nonumber  \\
&  \mbox{\scriptsize$ = \ds\sum_{x^r} P(Y_{t+1} \cond X_{t+1}=x) P(X_{t+1}=x, X_t=x^r, Y_{0:t}) $}  \nonumber  \\
&  \mbox{\scriptsize$ =  \mbox{\scriptsize $\ds\sum_{x^r} P(Y_{t+1} \cond X_{t+1}=x) P(X_{t+1}=x, X_t=x^r)  P(X_t = x^r, Y_{0:t}) $} $}  \nonumber  \\
&  \mbox{\scriptsize$ \propto \ds\sum_{x^r} \epsilon_\theta(x_{t+1}, Y_{t+1}), \tau_\theta(x^r, x) \phi_t(x^r) $} 
\label{eq:filter_recursion}
\end{align}

\subsubsection{Auxiliary function recursion}
%


Given the filter $\phi_{t} (x)$, the  {\bf backwards retrospective probability} can be computed as
\begin{align}
r_t(k^r \cond k) &:= P(X_t = k^r \cond X_{t+1} = k, Y_{0:n})   \nonumber \\
& \propto \phi_{t, \nu, \theta}(k^r) \tau_\theta (k^r, k) 
\end{align}
The auxiliary variable recursion can be given in terms of that quantity:
\begin{equation}
\label{eq:auxilliary_recursion}
\widehat{\rho}_{t+1}(k) \propto \sum_{k'} \bigg( \gamma_{t+1} s(k',k, Y_{t+1}) + (1-\gamma_{t+1}) \widehat{\rho}_t(k')  \bigg) r_t(k' \cond k)
\end{equation}
where $(\gamma_t)_{t \geq 1}$ is a decreasing sequence of step-sizes satisfying $\sum_{t \geq 1} \gamma_t = \infty$ and $\sum_{t \geq 1} \gamma_t^2 < \infty$.


\subsubsection{Decomposition}

Assuming the standard scenario where the state variables $(x_t)$ are discrete and $\theta$ is separable (into the state transition matrix $\tau$ and emissions distributions $\epsilon_k$), we may decompose the auxiliary function.   The {\bf transition auxilliary function} is 
\begin{equation}
\label{eq:auxiliary_function_tau}
\mbox{ \scriptsize $\widehat{\rho}^\tau_{t, \nu, \theta}(i, j, k ; \theta)= \df{1}{t} \mathbb{E}_{\nu, \theta} \big[ \ds\sum_{r=0}^t \mathds{1}\{ X_{r-1} = i, X_r =j  \} \cond Y_{0:t}, X_t=k \big]$ } 
\end{equation}
and the recursion (\ref{eq:auxilliary_recursion}) simplifies to 
\begin{equation}
\label{eq:rho_tau_recursion}
\mbox{ \scriptsize $\widehat{\rho}^\tau_{t+1}(i,j,k) = \gamma_{t+1} \one{j=k} r(i \cond j) + (1-\gamma_{t+1}) \ds\sum_{k'=1}^m \widehat{\rho}^\tau_t(i,j,k') r_{t+1}(k' \cond k)$ }
\end{equation}
This quantity is a $K \times K \times K$ matrix, and the $(i,j,k)$th entry provides the expected number of (latent=i, latent=j) bigrams for a single "observation" of a bigram given that the final latent state is $k$. 

Under the same scenario, the {\bf emissions auxilliary function} is 
\begin{equation}
\label{eq:auxiliary_function_epsilon}
\mbox{ \scriptsize $ \widehat{\rho}^\epsilon_{t, \nu, \theta}(i,k ; \theta)= \df{1}{t} \mathbb{E}_{\nu, \theta} \big[ \ds\sum_{r=0}^t \mathds{1}\{ X_{r} = i, s(Y_r) \}   \cond Y_{0:t}, X_t=k \big]  $}  \\
\end{equation}
and the recursion (\ref{eq:auxilliary_recursion}) simplifies to 
\begin{equation}
\label{eq:rho_epsilon_recursion}
\mbox{ \scriptsize $\widehat{\rho}^\epsilon_{t+1}(i,k) = \gamma_{t+1} \one{i=k} s(y_{t+1}) + (1-\gamma_{t+1}) \ds\sum_{k'=1}^m \widehat{\rho}^\epsilon_t(i,k') r_{t+1}(k' \cond k)$ }
\end{equation}
where $s(\cdot)$ are the sufficient statistics for the emissions distribution.  For example, $s(Y_t) = \one{Y_t = i}$ for categorical emissions and $s(Y_t) = (Y_t, Y_t Y_t')$ for normal emissions. In the context of categorical emissions, we may render the emissions auxiliary variable as a $K \times W \times K$ matrix, where the $(i,w,k)$th entry provides the expected number of (latent=i, observed=w) bigrams for a single "observation" of a bigram given that the final latent state is $k$.



\subsubsection{E-step}
Using the decomposition of the auxiliary function given by (\ref{eq:auxiliary_function_tau}) and (\ref{eq:auxiliary_function_epsilon}), we may decompose (\ref{ESS_current}) to obtain current estimates for the expected complete-data {\bf transition statistics} as:
\begin{equation}
\label{estimand_decomposed}
\widehat{S}^\tau_{t+1}(i,j)=   \ds\sum_{k=1}^K \widehat{\rho}^\tau_{t+1}(i,j,k) \widehat{\phi}_{t+1}(k)
\end{equation}
and current estimates for the expected complete-data {\bf emissions statistics} as
\begin{equation}
\label{estimand_decomposed}
\widehat{S}^\epsilon_{t+1}(i)=   \ds\sum_{k=1}^K \widehat{\rho}^\epsilon_{t+1}(i,k) \widehat{\phi}_{t+1}(k)
\end{equation}

\subsubsection{M-step}

Parameter optimization is immediate given the computation of $(\widehat{S}^\tau, \widehat{S}^\epsilon)$  For instance, the update to the state transition matrix is given by $\widehat{\tau}_{ij} = \widehat{S}^\tau_t(i,j)  / \sum_{j} \widehat{S}^\tau_t(i,j)$.  The update to the $k$th emissions distribution in the case of an HMM on categorical observations is $\widehat{\lambda_k} \propto S^\epsilon_t(k)$.

\section{Streaming training for \name} \label{AlgSec}

Algorithm \ref{alg} provides a psuedo-code implementation of \name.\footnote{A pip installable python package is also available, and will be provided in final form at publication time or can be provided now in draft form upon request.}  The algorithm is presented in a high-level way, merely assuming access to an EM-based streaming HMM algorithm.\footnote{In particular, one need not specifically implement the EM-based streaming HMM algorithm of \cite{cappe}; one could choose others \cite{khreich}.} This allows us to present the psuedo-code in a way that offloads description of things like recursion variables and step-size parameters to the streaming HMM algorithm itself.\footnote{Compared to other classes of streaming algorithms for estimating HMM parameters, the EM-based streaming algorithms provide more accurate state estimates and are easier to implement \cite{khreich}.  However, other techniques (e.g., gradient-based and minimum prediction error) may be more appropriate when $K$ is large.  We leave the application of \name to those classes of streaming HMM algorithms the reader.}  

\begin{algorithm}[tbh]
\begin{algorithmic}
  \Require Black-box predictive probability distributions $ \{ f^k\}_{k \in \blackbox}$   
  \Require Initialized \name model parameters, $\theta$
  \Require \# hidden states: $K : K >= | \blackbox|$
  \Require Streaming HMM algorithm based on EM 
    \Require $t_\text{min}$, warm-up period before M-step updates 
  \\
  \For{$t \in \{1,2,3,...\}$}
    \State{Update black-box predictive probability distributions, $\epsilon(y_t \cond k)  = f^k_{y_{1:t}}(\cdot)$ for $k \in \blackbox$ }
    \State{Do E-step for $k \in [1:K]$ }
   \If{$t \geq t_\text{min}$} 
       \State{Do M-step on $ \theta= \big( \pi, \tau, \{ \lambda_k\}_{k \in  \perturbed} \big) $}
   \EndIf
  \EndFor \\
  \Return \name model parameters, $\theta$
\end{algorithmic}
\caption{Perturbed sequence modeling using \name 
\label{alg}}
\end{algorithm}

It may seem surprising that Algorithm \ref{alg} could learn the probabilistic perturbations around the fixed predictive probability distributions $\{ f^k\}_{k \in \blackbox}$  from the source models $\{\mathcal{M}_k\}_{k \in \blackbox}$, given that the source models  provide \name with a stochastic process of emissions parameters, $\{  \lambda_k^t\}_{k \in \source, t \geq 1}$ (and therefore a stochastic process of emissions distributions, $\{  \epsilon_k^t\}_{k \in \source, t \geq 1}$), rather than the constant emissions distributions of a standard HMM.  On the surface, this might seem to complicate the learning of the remaining (perturbation) emission distributions,  $\{  \epsilon_k\}_{k \in \perturbation}$, as well as the transition dynamics across this heterogenous system.

However, as it turns out, the streaming HMM algorithm readily wraps around the source predictive probability functions, so long as they are kept updated at each iteration of the streaming algorithm.  To see this, simply consider the ingredients of the streaming recursions from the online HMM updates:
\begin{itemize}
\item The filter recursion (\ref{eq:filter_recursion}) depends upon the recursion variable itself, the state transition matrix $\tau$ and emissions distributions $\{\epsilon_k\}_{k \in \mathcal{K}}$.

\item The transition auxiliary variable recursion (\ref{eq:rho_tau_recursion}) depends upon the recursion variable itself, the step-size hyperparameter, and the transition parameter $\tau$. 

\item The emissions auxiliary variable recursion (\ref{eq:rho_epsilon_recursion}) depends upon the recursion variable itself, the step-size hyperparameter, the transition parameter $\tau$, and the sufficient statistics for the emissions distribution computed from an observed individual datum. 
\end{itemize}

Access to these quantities are available at any timestep $t$.   In particular:
\begin{itemize}
\item $\{\epsilon_k\}_{k \in \mathcal{K}}$ are available from the M-step for $k \in \perturbed$ and from provided scoring functions $\{ f^k\}_{k \in \blackbox}$ for  $k \in \pretrained$.  
\item All entries of $\tau = (k,k')_{k,k' \in  \mathcal{K}}$ are available due to the M-step. 
\end{itemize}
 Thus, so long as the black-box predictive probability distributions $ \{ f^k\}_{k \in \blackbox}$ are pre-trained rather than learned, we can apply Algorithm \ref{alg} to learn the parameters $\theta$ of \name. 
 
\section{Experiment 1 (Simulation):  Are \name estimators consistent?}

Roughly speaking, an estimator is {\bf consistent} if the estimator applied to the whole population equals the parameter \cite{martin}.  In other words, as more samples are taken, the estimator will converge to the true value of the parameter.    It can be shown that the maximum likelihood estimator for a HMM's $\theta$ parameters is strongly consistent (i.e. converges \emph{a.s.}) under a minimal set up assumptions \cite{douc}.\footnote{Although it seems that, for batch EM training, statistical consistency can be achieved only by using more batch EM iterations as the sample size grows\cite{cappe}.}     The estimator obtained from the streaming training algorithm of \cite{cappe} is not as well-understood analytically, but its consistency is well-supported by a heuristic analytical argument and with empirical simulations \cite{cappe}.    But is it reasonable to expect the same behavior with \name?

We conjecture that the answer is yes, because the proof of consistency for the maximum likelihood estimator of HMM parameters (and the heuristic proof for the corresponding streaming estimators of \cite{cappe}) would carry over to \name, given the arguments of Section \ref{AlgSec}.  In other words, both the algorithmic recursions and the proof should be \quotes{blind} to the fact that some emissions are pinned to an external stochastic process. 

Here we empirically evaluate whether the estimators of Algorithm \ref{alg} are consistent for the \name parameters $\theta$.

\subsection{Methodology}

To evaluate consistency (which depends on convergence) we need to have a notion of distance between models.  Luckily, \cite{juang} provides a notion of divergence between HMM model:
\begin{align*}
D(\theta_0, \theta) = \ds\lim_{t \to \infty} D_t = \ds\lim_{t \to \infty} \ds\frac{1}{t} \bigg[ \log  P(Y_t \cond \theta_0) -\log P(Y_t \cond \theta) \bigg]
\end{align*}
In practice, one recursively generates samples $Y_{1:t} \sim \text{HMM}(\theta_0)$ for increasing $t \geq1$ and computes $D_t$ until $\{ D_s\}_{s=1}^t$ meets a convergence threshold.  We then approximate $D(\theta_0, \theta)  \approx D_t$.  We can co-opt $D$ as a distance\footnote{Technically, this isn't a mathematical distance function, or metric, due to the function's non-symmetry.  However, this function is easily symmetrizable but taking the mean of $D(\theta_0, \theta)$ and $D(\theta, \theta_0)$.} between \name models by applying it to the PROPS parameters instead. 

To check if Algorithm \ref{alg} provides consistent estimates of the \name model, we sample $Y_{1:t} \sim \name(\theta_0)$ for some $t$, and then we apply Algorithm \ref{alg} to form estimator $\widehat{\theta}_t$.     If the \name estimator is consistent, then we should have that $D(\theta_0, \widehat{\theta}_t) \to 0$ as $t$ increases. 

We evaluate consistency for a \name model with $K=3, \numbb=1, f_t \stackrel{i.i.d}{\sim}$ Dirichlet($\one{W})$, categorical emissions, and various $W$.\footnote{In particular, the choice of $f_t$ reflects the fact that a black-box feedforward model is sufficiently complex that its sequence of predictive probability distributions may look like independent draws from a Dirichlet.}   Note that computing $D(\theta_0, \widehat{\theta}_t)$ between a true and an estimated model requires a \emph{separate} hold-out sample, $Y'_{1:t} \sim \name(\theta_0)$.  Each time a sample is drawn from $\name(\theta_0)$, we perturb $f_t$ slightly by drawing $f_t' \stackrel{i.i.d}{\sim} \text{Dirichlet}(\alpha \, f_t)$ for concentration parameter $\alpha$=100.

\subsection{Results and Discussion}

\begin{figure}
  \includegraphics[width=0.5\textwidth]{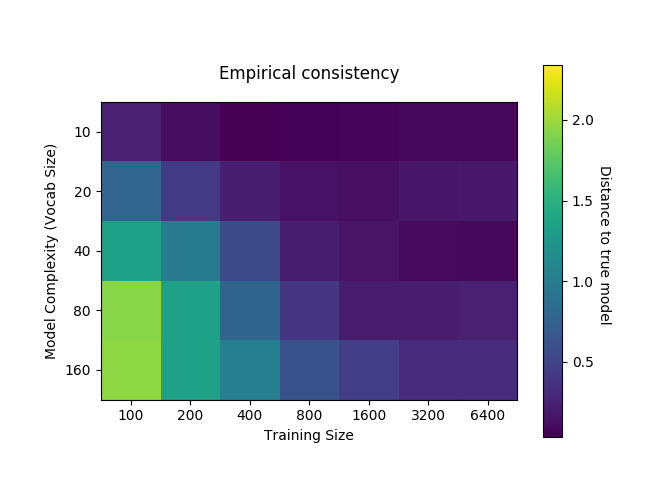}
  \caption{ \emph{Empirical investigation on the consistency of the \name estimator provided by Algorithm \ref{alg}. }}
  \label{img:empirical_consistency}
\end{figure}

 Figure \ref{img:empirical_consistency} shows that, as desired, $D(\theta_0, \widehat{\theta}_t)$ decreases as sample size $t$ grows.  As expected, the rate of convergence is slower for models that are more complex. 


\section{Experiment 2: Probabilistically Perturbing a Wikipedia language model into a model of President Donald J. Trump's tweeting}

Now we investigate the utility of the \name model in the context of personalized language modeling.  

\subsection{Methodology}

The \rnn model was created on a g2.8xlarge EC2 instance (which is backed by four NVIDIA GPUs) using these steps:   

\begin{enumerate}
\item The entire Wikipedia corpus (2,581,793 articles, approximately 15GB on disk) was obtained using the {\tt gensim} python module.   There were approximately 2,568 words per article.  We removed punctuation and capitalization from the articles.  

\item A word2vec model was built on this corpus using the python module {\tt henson} with a random subcorpus of  5,000 Wikipedia articles.  In particular, a continuous bag of words (CBOW) model with a context window of 5 was used.    The embedding was chosen to have dimensionality 100.   The word had to appear at least 12 times in the subcorpus in order to survive.  This led to a vocabulary of $W=13,711$ words.  All other words were rendered as 'OOV'  (out of vocabulary). 

\item A stacked Long Short Term Memory (LSTM) network was then trained on the full Wikipedia corpus. The structure of the LSTM was one embedding layer (with embedding weights from the pretrained word2vec model) and two LSTM layers stacked together. The recurrent activation (for the input/forget/output gates) used was hard-sigmoid, whereas the activation used for the cell state and the hidden state was 'tanh'. Each LSTM cell has a output, and then the output from the whole LSTM layer is a list.  Each member of this list is fed into a dense layer, where the activation is the 'tanh.'   That output was fed into another dense layer, where the activation is softmax.  The RMSprop optimizer was used for training.  The loss was the categorical cross-entropy loss between the input sequence and the output sequence, where the output sequence is a shift left by one word of the input sequence. 
We treat 50 words as a sequence, and use a minibatch size of 5000 sequences. We trained for 2 epochs (i.e. went through the corpus two times).   The LSTM training took > 1 day. 
\end{enumerate}

Our target corpus for transfer learning was a corpus of tweets posted by President Donald J. Trump from 07/15/2014 until 12/21/2016.\footnote{Corpus obtained from \url{https://www.kaggle.com/austinvernsonger/donaldtrumptweets/\#data.csv}}   We removed urls, phrases with leading hashtags, and punctuation. 

We used \name to perturb the baseline, black-box Wikipedia model into a probabilistic, personalized language model of Trump's tweeting.  The \name model was fit with $K=3, |\blackbox| = 1, W=13,711$.  The model was very lightly trained -- on the first 2,000 words in the tweet corpus, fed to the model in chronological order.   

For comparison, we also fit a standard streaming HMM model to the tweets alone.  The HMM model was fit with  $K=3, W=13,711$.



\subsection{Results and Discussion}

\begin{figure}
  \includegraphics[width=0.5\textwidth]{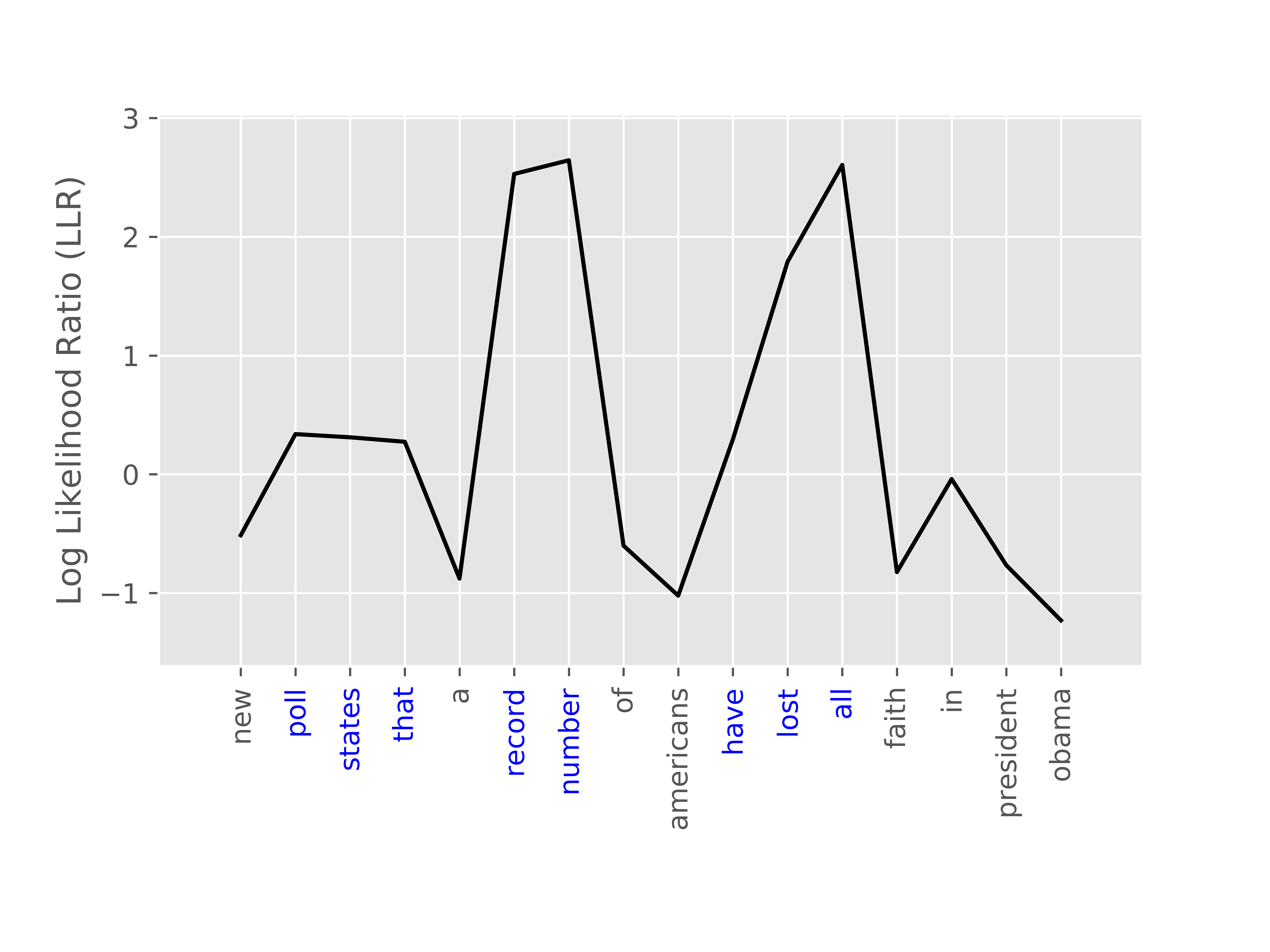}
  \caption{ \emph{Log likelihood ratio of predictive probability scores  for the \name model (trained on Trump tweets) versus a standard HMM model (trained on Trump tweets).}  Words with large positive values are colored in blue.}
  \label{img:v_hmm}
\end{figure}

In Figure \ref{img:v_hmm}, we show the log likelihood ratio of {\bf predictive probability scores} (i.e. $P(y_t \cond y_{1:t-1}) = \sum_k P(y_t \cond x_t =k) P(x_t =k \cond y_{1:t-1})$) for the \name model versus a standard HMM model on a representative subtweet.   We first note that the mean is positive, meaning that the \name model outperforms a standard HMM model in predicting the next word.   Indeed, the mean predictive probability is 0.01 for the \name model, compared to 0.004 for the standard HMM model (and $7.2 \cdot 10^{-5}$ for random guessing).   Further insight can be obtained by investigating the blue stretches of text -- i.e. the particular locations where the \name model outperforms the standard HMM.   The black-box predictive model is able to contribute knowledge about collocations that inhere in the English language generally ("poll states that", "a record number", "have lost all").  A standard HMM, trained from scratch on Donald J. Trump's tweets, would have virtually zero knowledge of such general collocations, particularly after a short training time. 

\begin{figure}
  \includegraphics[width=0.5\textwidth]{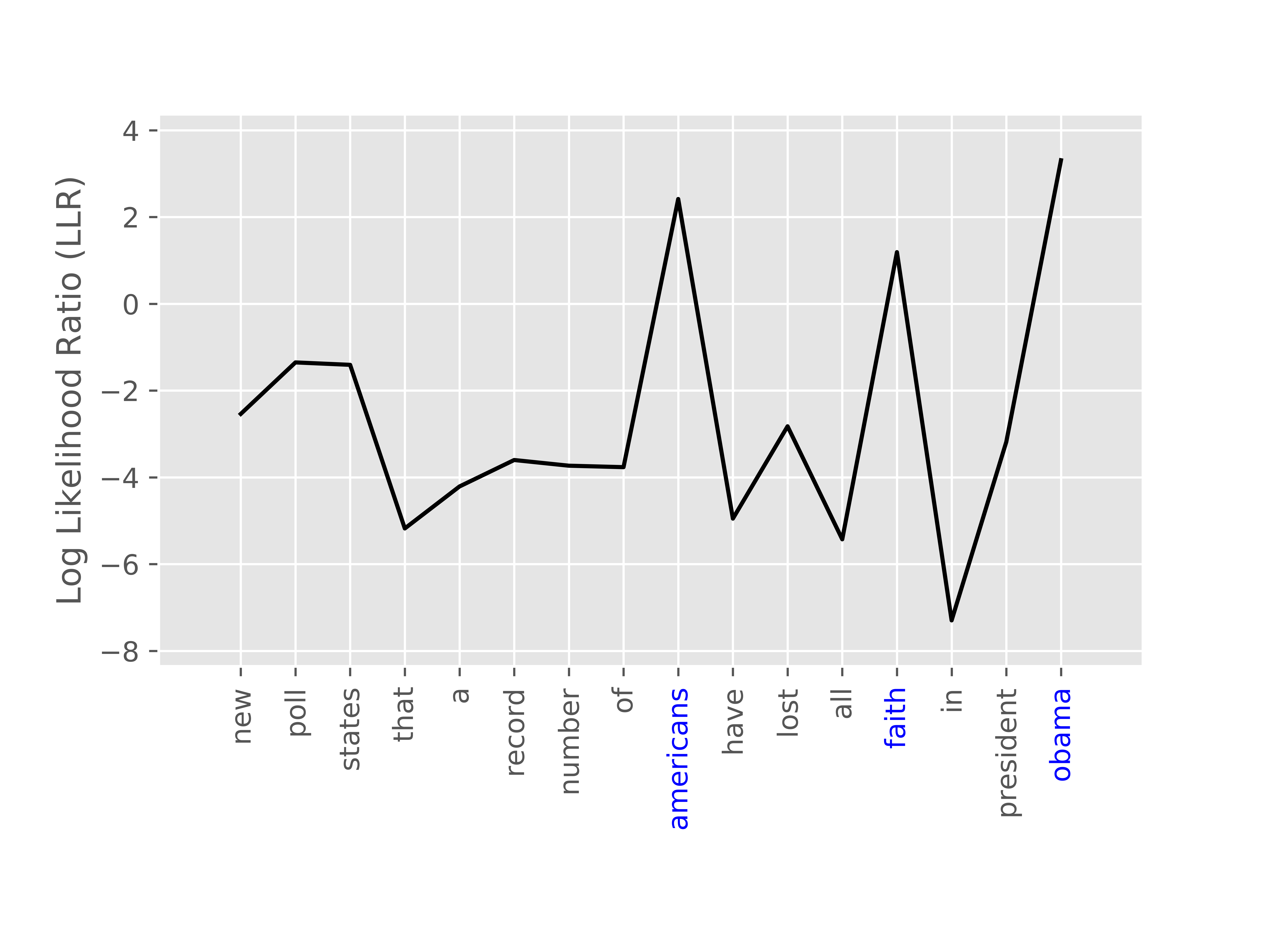}
  \caption{\emph{Log likelihood ratio for the \name model that the latent state underlying the observed language belongs to a local perturbation (Trump) mode rather than a baseline (Wikipedia) mode.} Words with large positive values are colored in blue.}
  \label{img:personalization}
\end{figure}

In Figure \ref{img:personalization}, we show the log likelihood ratio for whether the latent state has membership in a local perturbation mode.  In other words, if $p_t:=P(x_t \in \perturbed \cond y_{1:t})$, then $1-p_t:=P(x_t \in \blackbox \cond y_{1:t})$, and  we plot the log of $\frac{p_t}{1-p_t}$.    What we see here is that the \name model has successfully customized (or personalized) the more generic language model to handle the particular content of President Trump's tweeting.   The final word in the phrases "a record number of...Americans", "have lost all...faith", and "in president...Obama" are all substantially more likely to be observed in language from President Trump than in a broader corpus of English language (i.e. Wikipedia).    The local perturbation modes, $\perturbed$, of the \name model provide this capacity for personalization. 

\begin{figure}[htp]
  \includegraphics[width=0.5\textwidth]{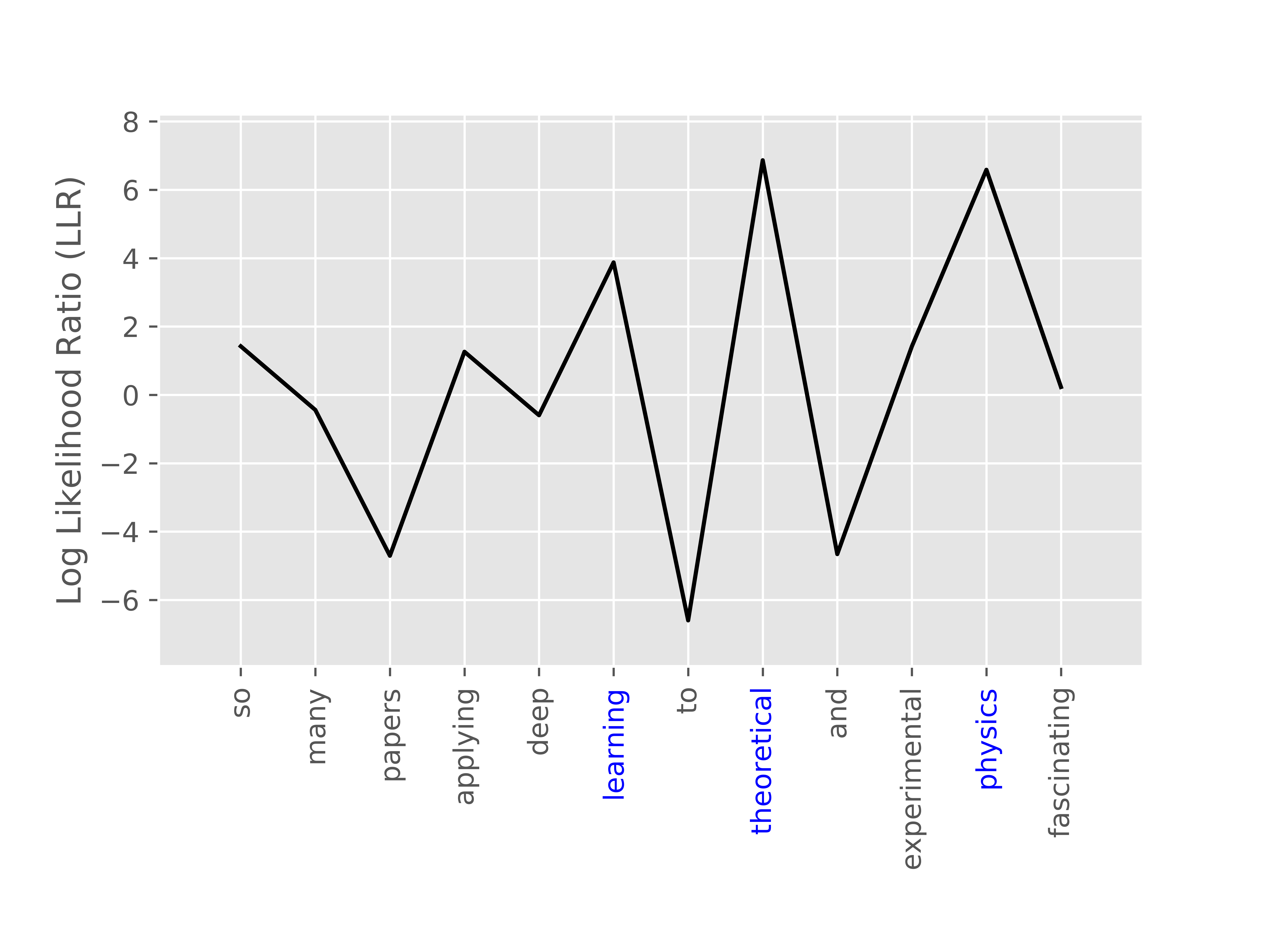}
  \caption{ \emph{Log likelihood ratio of predictive probability scores for the generic English language model versus a corresponding \name model that has been customized to President Donald J. Trump's tweets.}   The test data here is a tweet from machine learning researcher Yann Lecun. Words with large positive values are colored in blue.}
  \label{img:lecun}
\end{figure}

Of course, the cost of personalization is a poorer fit to the language of others.  In Figure \ref{img:lecun}, we show the log likelihood ratio of predictive probability scores for the \name model versus the baseline RNN model on a tweet from Yann LeCun on Oct. 8, 2018.\footnote{"So many papers applying deep learning to theoretical and experimental physics! Fascinating."}  Compared to the more generic RNN model, the \name model, which was personalized to Trump, was much more surprised by the ending of phrases like  "to...theoretical",  "and....experimental physics", and "applying deep...learning."   

We emphasize the lightweight nature of the transfer.  Whereas the source RNN model was trained on $\sim$ 6.6 billion words from the Wikipedia corpus, the \name customization was performed on mere 2,000 additional words from Trump tweets. 

\section{Related Work}

The (RNN-based) language model personalization scheme of \cite{yoon} provides many of the desiderata outlined in the introduction.  By forming a source RNN on a large dataset and then retraining only the last layer on the endpoint, they obtain a lightweight, fast transfer learning scheme for sequential data that respects user privacy.    The primary drawback of \cite{yoon}  relative to \name is the loss of interpretability in the personalized model.  Because the hidden variables of that model are not stochastic, one loses any insight into the respective contributions of the original source model vs. the personalization modes towards predictions about upcoming behavior.   A subsidiary drawback is the loss of reliability as the more expressive model will also have higher variance. 

A \emph{stochastic recurrent neural network} (SRNN) \cite{fraccaro} addresses a shortcoming of RNN's relative to state-space models such as HMM's by allowing for stochastic hidden variables.  We surmise that the the SRNN framework will eventually generate a state-of-the-art transfer learning mechanism for sequential data that satisfies the interpretability desideratum from the introduction.  However, to our knowledge, such a mechanism has not yet been developed.  Moreover, the training of a SRNN is substantially more complex than training a standard RNN, let alone a HMM, and one would expect that computational complexity to spill over into the transference algorithm.   If so, \name would provide a lightweight alternative. 

\section{Conclusion}

The \name model is a lightweight transfer learning mechanism for sequence learning which learns probabilistic perturbations around the predictions of one or more arbitrarily complex, pre-trained black box models.   In the results section, we saw that the \name model has an advantage over a standard HMM, because it has a \quotes{head start} -- for example, it knows about collocations in the English language before ever seeing a single Trump tweet.   On the other hand, the \name model has an advantage over a generic black-box RNN model, because it is customized to the target audience.   While there are other schemes for transfer learning with RNN's, \name is a fully probabilistic model.  In particular, one can perform inference (filtering, smoothing, and MAP estimates) on the hidden states.    With this inference, one can make probability statements about when an agent is behaving like the general corpus, when they are behaving in ways that are more characteristic of themselves, and when they are acting unusually more generally.  

\bibliography{references}{}
\bibliographystyle{ieeetr}


%
%


\end{document}